\documentclass[letterpaper,10pt,conference]{ieeeconf}
\IEEEoverridecommandlockouts
\let\labelindent\relax
\usepackage{graphicx} \graphicspath{ {figures/} }
\usepackage{amsmath,amssymb}
\usepackage{algorithmic}
\usepackage[ruled]{algorithm2e}
\usepackage{acronym}
\usepackage{enumitem}
\usepackage{booktabs}
\usepackage{hyperref}
\usepackage{balance}
\usepackage{xspace,setspace}
\usepackage[skip=3pt,font=small]{subcaption}
\usepackage[skip=3pt,font=small]{caption}
\usepackage[dvipsnames]{xcolor}
\usepackage[capitalise]{cleveref}
\usepackage{tabularx,colortbl,multirow,array,makecell}
\usepackage{overpic}
\usepackage{cite}
\usepackage{tikz}
\usepackage[T1]{fontenc}
\usepackage{mathptmx}

\makeatletter
\DeclareRobustCommand\onedot{\futurelet\@let@token\@onedot}
\def\@onedot{\ifx\@let@token.\else.\null\fi\xspace}
 
\def\ie{\emph{i.e}\onedot}

\def\etal{\emph{et al}\onedot}
\makeatother

\crefname{algocf}{alg.}{algs.}
\Crefname{algocf}{Algorithm}{Algorithms}

\makeatletter
\def\BState{\State\hskip-\ALG@thistlm}
\makeatother

\makeatletter
\renewcommand{\paragraph}{%
  \@startsection{paragraph}{4}%
  {\z@}{0ex \@plus 0ex \@minus 0ex}{-1em}%
  {\hskip\parindent\normalfont\normalsize\bfseries}%
}
\makeatother

\crefname{algocf}{alg.}{algs.}
\Crefname{algocf}{Algorithm}{Algorithms}

\definecolor{gblue}{HTML}{4285F4}
\definecolor{gred}{HTML}{DB4437}

\acrodef{dof}[DoF]{Degree of Freedom}
\acrodef{vbts}[VBTS]{Vision-based Tactile Sensors}
\acrodef{mbts}[MBTS]{Magnetic-based Tactile Sensors}
\acrodef{sr}[SR]{Super-Resolution}
\acrodef{hr}[HR]{High-Resolution}
\acrodef{lr}[LR]{Low-Resolution}
\acrodef{vae}[VAE]{Variational Auto-encoder}
\acrodef{cvae}[CVAE]{conditional variational autoencoder}
\acrodef{ddpm}[DDPM]{Denoising Diffusion Probabilistic Model}
\acrodef{cnn}[CNN]{Convolutional Neural Network}
\acrodef{mlp}[MLP]{Multilayer Perceptron}
\acrodef{gan}[GAN]{Generative Adversarial Networks}
\acrodef{knn}[k-NN]{k-nearest neighbors}
\acrodef{fc}[FC]{fully-connected}
\acrodef{psnr}[PSNR]{Peak-Signal-to-Noise Ratio}
\acrodef{ssim}[SSIM]{Structural Similarity}
\acrodef{fid}[FID]{Frechet Inception Distance}

\newcommand{\name}{SuperMag\xspace}

\title{\LARGE \bf SuperMag: Vision-based Tactile Data Guided High-resolution Tactile Shape Reconstruction for Magnetic Tactile Sensors}

\author{Peiyao Hou$^{1,2\star}$, Danning Sun$^{1\star}$, Meng Wang$^{2}$, Yuzhe Huang$^{1,2}$,\\ Zeyu Zhang$^{2}$, Hangxin Liu$^{2}$, Wanlin Li$^{2\dagger}$, Ziyuan Jiao$^{2\dagger}$
\thanks{This work was supported in part by the National Natural Science Foundation of China (Grant No.52305007), and by the State Key Laboratory of Mechanical System and Vibration (Grant No. MSV202519).}%
\thanks{$^{\star}$ Peiyao Hou and Danning Sun contributed equally to this work. This work was conducted during Peiyao Hou's internship at the Beijing Institute for General Artificial Intelligence (BIGAI). $^\dagger$~Corresponding authors. $^{1}$~Department of Automation, Beihang University. $^{2}$~State Key Laboratory of General Artificial Intelligence, BIGAI, Beijing, China.}%
}%
\begin{document}

\maketitle
\thispagestyle{empty}
\pagestyle{empty}

\begin{abstract}
Magnetic-based tactile sensors (MBTS) combine the advantages of compact design and high-frequency operation but suffer from limited spatial resolution due to their sparse taxel arrays. This paper proposes SuperMag, a tactile shape reconstruction method that addresses this limitation by leveraging high-resolution vision-based tactile sensor (VBTS) data to supervise MBTS super-resolution. Co-designed, open-source VBTS and MBTS with identical contact modules enable synchronized data collection of high-resolution shapes and magnetic signals via a symmetric calibration setup. We frame tactile shape reconstruction as a conditional generative problem, employing a conditional variational auto-encoder to infer high-resolution shapes from low-resolution MBTS inputs. The MBTS achieves a sampling frequency of 125~Hz, whereas the shape reconstruction sustains an inference time within 2.5~ms. This cross-modality synergy advances tactile perception of the MBTS, potentially unlocking its new capabilities in high-precision robotic tasks.
\end{abstract}

\section{Introduction}
Tactile sensing is essential in robotics, enabling agents to perceive and interact with their environment through physical contact~\cite{cutkosky2008force,yousef2011tactile}. 
Inspired by the biological sense of touch, tactile sensors detect mechanical stimuli such as contact force, texture, slip, and vibrations. Common sensing technologies include capacitive~\cite{dawood2023learning}, resistive~\cite{zhu2022recent}, piezoresistive~\cite{stassi2014flexible}, piezoelectric~\cite{lin2021skin}, triboelectric~\cite{qu2022fingerprint}, barometric~\cite{hou2024location}, optical~\cite{yao2024recent}, and magnetic~\cite{man2022recent} sensors, each offering unique advantages for tactile perception and robotic applications. In general, a high-quality tactile sensor should: i)~have a conformal contact surface for effective grasping and manipulation; ii)~be low-cost, easy to fabricate, and ideally open-source; iii)~feature a compact design for use in confined environments; and iv)~detect diverse contact information, such as multi-axis forces and shape reconstruction, with higher spatial resolution improving performance in precise robotic tasks.

Among these sensing techniques, \ac{mbts}~\cite{wang2016design} offer advantages such as compact and simple designs, high response frequencies ($>100$~Hz), multi-axis force detection, and cost-effectiveness. A key limitation of \ac{mbts}, shared with other non-vision-based methods, is their taxel-array configuration, which restricts spatial resolution due to the physical space occupied by each sensing element. This limitation impedes their performance in applications requiring fine-grained tactile perception.

To address this limitation, \ac{sr} techniques for \ac{mbts} have been studied to enhance spatial resolution beyond sparse taxel arrays, enabling detailed texture and shape perception~\cite{sun2022guiding}. Current approaches often use indentation probes on multi-axis linear stages to collect high-resolution data via controlled tapping motions~\cite{huang2024multipole,yan2021soft,hu2024large}, but they are limited to discrete contact points or simple multi-tap interactions, restricting their use for complex, continuous surfaces. Recent reconstruction-based methods~\cite{wu2022tactile} show promise in generating high-resolution tactile shapes from \ac{mbts} readings; however, they rely on complex setups and conversions to acquire high-resolution supervisory signals, and struggle to generalize to unseen objects. 

\begin{figure}[t!]
    \centering
    \includegraphics[width=\linewidth]{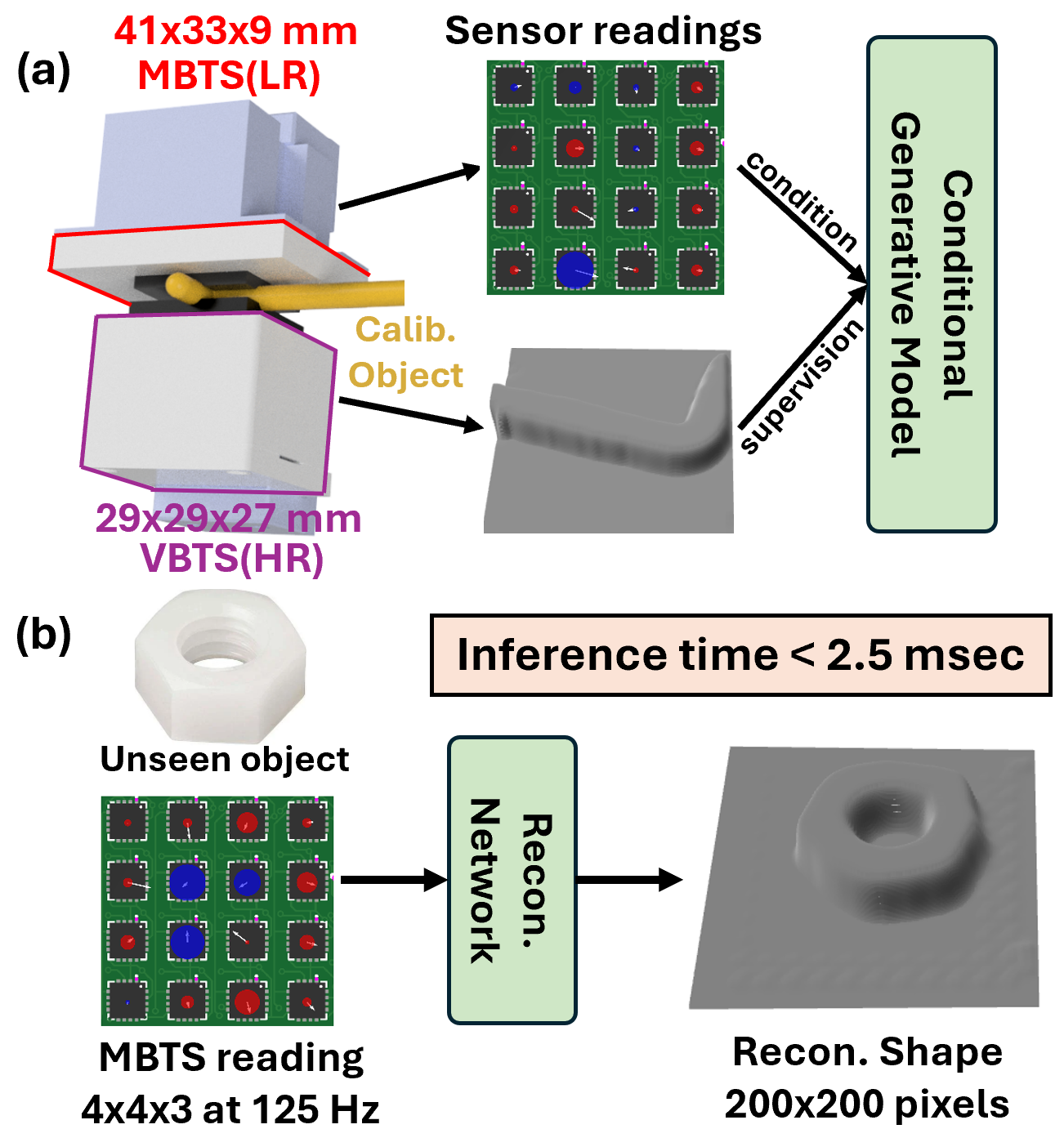}
    \caption{\textbf{SuperMag: High-resolution Tactile Shape Reconstruction for \acf{mbts} with \acf{vbts} data.} (a)~Training: \acf{hr} \acs{vbts} depth images serve as supervisory signals to guide \acf{lr} \ac{mbts} in reconstructing high-resolution tactile shapes of the object. (b)~Inference: Sparse \ac{mbts} data are used to reconstruct the tactile shape of an unseen test object.}
    \label{fig:teaser}
\end{figure}

Recent advancements in optical methods, particularly \ac{vbts}~\cite{shah2021design}, offer a promising solution to these limitations. Despite their bulky form factor and lower response frequencies (30-60~Hz), \ac{vbts} inherently achieve high-resolution shape reconstruction through direct visual feedback, even for unseen, intricate textures, by leveraging learned pixel-level depth mappings from minimal training data~\cite{zhong2023touching,li2024vision}. The complementary strengths and weaknesses of \ac{mbts} and \ac{vbts} suggest a synergistic potential. We propose that the easily acquired tactile data from \ac{vbts}---which encapsulate fine geometric and textural details---could serve as supervisory signals to guide \ac{mbts} in reconstructing high-resolution shapes. By integrating cross-modal learning frameworks, the high-resolution priors captured by \ac{vbts} could enable \ac{mbts} to surpass their physical resolution limits, bridging the gap between sparse tactile data acquisition and dense, accurate shape estimation. 

In this work, we propose \name, a tactile shape reconstruction method that leverages high-resolution tactile data from \ac{vbts} to guide the super-resolution of low-resolution \ac{mbts} signals, enabling high-spatial-resolution shape reconstruction at high operational frequencies. 
To achieve this, we design and fabricate open-source, low-cost \ac{vbts} and \ac{mbts} with identical contact modules for compatibility~\cite{bhirangi2021reskin,lin2023dtact}.
A symmetric calibration setup (see \cref{fig:teaser}(a)) enables seamless co-located collection of low-resolution magnetic data and high-resolution visual-depth ground truth.
We frame the reconstruction task as a conditional generative problem, where high-resolution depth maps are inferred from sparse \ac{mbts} readings using a \ac{cvae}. 
Experiments show that \name outperforms baseline methods in both quantitative metrics and qualitative evaluations, achieving high-resolution shape reconstruction with an inference time of 2.5~ms per reading. This synergy between vision-based and magnetic-based modalities presents a novel, unexplored avenue to advance tactile super-resolution, potentially unlocking new capabilities in robotics for tasks requiring precise tactile perception using \ac{mbts}.

\section{Related Work}
\subsection{\acf{mbts}}
\ac{mbts} use embedded magnets and magnetic field sensors, such as Hall effect sensors~\cite{rehan2022soft} or magnetoresistive sensors~\cite{ge2019bimodal}, to detect changes in the magnetic field caused by external forces on a deformable contact module, enabling high-frequency, high-sensitivity force detection and shape reconstruction. For instance, MagOne~\cite{wang2016design} employs a moving least-squares approach to decouple magnetic signals into normal and shear forces, achieving high resolution and low hysteresis. Dai~\etal~\cite{dai2024split} introduces a split-type sensor with centripetal magnetization and a theoretical decoupling model, enabling wireless 3D force sensing. MTS~\cite{li2024tactile} utilizes a giant magnetoresistance sensor, achieving shear force sensing and pressure detection supported by theoretical deformation models. ReSkin~\cite{bhirangi2021reskin} leverages a low-cost magnetized elastomer skin and compact magnetometer-based sensing for distributed three-axis force measurement and contact localization. Collectively, \ac{mbts} provide a compact, low-cost, high-frequency solution for tactile sensing, but their limited spatial resolution remains challenging.

\subsection{\acf{sr}}
\ac{sr} techniques for enhancing tactile spatial resolution can be categorized into model-based and learning-based methods. Model-based approaches, such as interpolation~\cite{gribbon2004novel,hu2022wireless}, apply spatial smoothing to reconstruct high-resolution tactile data from low-resolution arrays, improving localization accuracy. For example, Lepora~\cite{lepora2015superresolution} leverages active Bayesian perception to achieve 40-fold \ac{sr} in object localization. Learning-based methods leverage data-driven models to infer sub-taxel details from high-resolution ground-truth data. For instance, soft magnetic skin~\cite{yan2021soft} achieves 60-fold \ac{sr} via \ac{mlp}, while Hu~\etal~\cite{hu2024large} combines \ac{knn} clustering and \ac{cnn} for 45-fold \ac{sr}. Huang~\etal~\cite{huang2024multipole} propose a three-layer \ac{mlp} for 70-fold \ac{sr} positioning, and Wu~\etal~\cite{wu2022tactile} use \ac{cnn} and \ac{gan} for 100-fold \ac{sr}. While these methods have shown promise, they are subject to certain limitations, including dependence on probing systems, the need for intricate setups to achieve high-resolution data collection, and challenges in generalizing effectively to previously unseen objects. In this work, we propose a novel \ac{sr} method that leverages \ac{vbts} to guide \ac{mbts} for complex texture reconstruction using simple training objects. By formulating the problem as a generative reconstruction problem, our approach enables \ac{sr} shape reconstruction from \ac{mbts}, revealing fine details not captured in previous work, using a simple data collection process.

\subsection{\acf{vbts}}
\ac{vbts} utilize cameras to capture deformations in elastomer, providing pixel-level tactile feedback. These sensors operate primarily through two approaches: marker-based tracking and reflection-based reconstruction. Marker-based methods use monocular~\cite{yamaguchi2017implementing} or binocular cameras~\cite{cui2021self} to track embedded markers, inferring distributed forces and contact locations. For instance, TacTip~\cite{ward2018tactip} tracks white-tipped pins in a hemispherical elastomer, while GelForce~\cite{sato2009finger} estimates forces by monitoring red and blue marker arrays. Reflection-based methods~\cite{wang2024large, li2024minitac} reconstruct contact surfaces by analyzing light field changes, and mapping intensity to surface gradients or depth. Examples include GelSight~\cite{yuan2017gelsight}, GelSlim~\cite{donlon2018gelslim}, and DIGIT~\cite{lambeta2020digit}, which use RGB lighting and Poisson solvers~\cite{mckenney1995fast} to compute depth maps. Markers can also be incorporated for force inference~\cite{yuan2015measurement, li20233, zhao2024tac}. DTact~\cite{lin2023dtact} directly computes depth maps from darkness variations in a semi-transparent elastomer under white light. While \ac{vbts} offer high-resolution tactile feedback, they face challenges such as bulky form factors, low operational frequencies, and heating issues from processing large image volumes during prolonged use. In this paper, we seek to leverage the high-resolution capabilities of \ac{vbts} to guide and enhance the spatial resolution of \ac{mbts}, enabling high-frequency, high-fidelity tactile perception while overcoming the limitations of each modality individually.

\begin{figure*}[t]
    \centering
    \includegraphics[width=\linewidth]{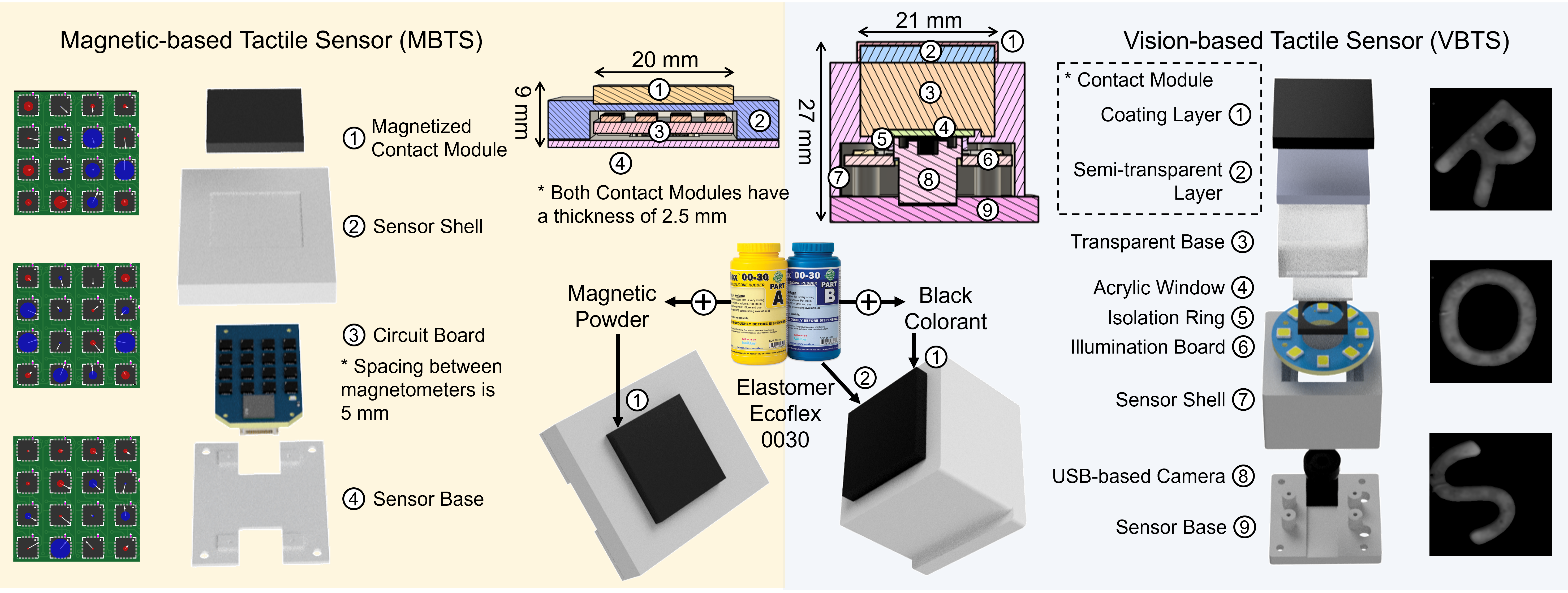}
    \caption{\textbf{Exploded and schematics views demonstrating the key components of the two tactile sensors examined in this work.} 
    (a)~The $9$ mm-thick \ac{mbts}~\cite{bhirangi2021reskin} features a magnetized contact module, mounted over a circuit board with $4\times4$ magnetometers spaced 5 mm apart.
    (b)~The $27$ mm-thick \ac{vbts}~\cite{lin2023dtact} uses a camera module embedded beneath an acrylic window and illumination ring, together with a contact module composed of a coating layer and a semi-transparent layer. Both contact modules are 2.5 mm thick, with a nominal width of 20 mm (\ac{mbts}) and 21 mm (\ac{vbts}), respectively.}
    \label{fig:sensor}
\end{figure*}

\section{Tactile Sensors} \label{sec:sensor}
This section outlines the data collection pipeline for \name, along with the criteria for selecting tactile sensors, their design, fabrication, and calibration procedures.

\subsection{Sensor Selection Criteria} 
The selection criteria for tactile sensors in our proposed framework are as follows:

\begin{enumerate}[label=\arabic*), leftmargin=*, noitemsep, nolistsep]
\item Both \ac{vbts} and \ac{mbts} must have contact modules with identical dimensions and materials to ensure consistent contact feedback during interactions.
\item Both sensors should be cost-effective, easy to customize, and simple to fabricate, ensuring reproducibility.
\item \ac{mbts} measures distributed multi-axis forces (via magnetic flux), \ac{vbts} provide high-resolution shape details.
\end{enumerate}

Based on these criteria, we adopt and customize~\cite {bhirangi2021reskin} as the \ac{mbts} and~\cite{lin2023dtact} as the \ac{vbts}. Below, we detail their design and fabrication.

\subsection{Magnetic-based Tactile Sensor}
\ac{mbts} detects changes in the magnetic field between a magnetized elastomer skin and a magnetometer. The contact module, made of silicone mixed with magnetic powder, is molded into custom shapes. We selected the dimensions of $20\times 20\times 2.5$mm to match \ac{vbts} (see \cref{fig:sensor}(a)).

The fabrication process involves:
\begin{enumerate}[label=\arabic*), leftmargin=*, noitemsep, nolistsep]
    \item Mixing magnetic powder (MQP-15-7) with platinum-catalyzed silicone (Ecoflex 00-30 from Smooth-On, with shore hardness 00-30) at a 3:1:1 ratio.
    \item Degassing the mixture for 2 minutes in a vacuum chamber and pouring it into a 3D-printed mold.
    \item Magnetizing the composite using cuboid permanent magnets ($100\times 50 \times 20$ mm) and curing it for 4 hours at room temperature.
    \item Attaching the magnetized silicone to the sensor shell using silicone adhesive (Sil-Poxy from Smooth-On).
\end{enumerate}

A $4\times4$ magnetometer array (MLX90393) with a spacing of 5 mm measures 3-axis magnetic flux changes. These magnetometers are accessed via a SPI bus by an onboard MCU (ESP32PICO), and the data are framed and forwarded to an upper host via UART. The interface also supports a serial connection to facilitate deployment on multi-sensors. The sensor operates at 125~Hz, with overall dimensions of $41\times 33\times 9$~mm and a total cost of less than \$40.

\subsection{Vision-based Tactile Sensor}
\ac{vbts} employs a camera to capture light reflection within a semitransparent elastomer layer during interactions. The depth of contact is directly mapped to the intensity of darkness in the captured image, enabling pixel-level shape reconstruction. The contact module, which is also made of Ecoflex 00-30, has slightly larger dimensions ($21\times 21\times 2.5$~mm) to account for cropping (see \cref{fig:sensor}(b)).

The fabrication process includes:
\begin{enumerate}[label=\arabic*), leftmargin=*, noitemsep, nolistsep]
    \item Molding the transparent base with PDMS (Sylgard 184 from DOW, with a shore hardness of 50A), degassing the mixture (base: catalyst = 10: 1) for 2 minutes, and curing for 24 hours at room temperature.
    \item Pouring platinum-catalyzed silicone (1:1 ratio, degassed for 2 minutes) over the PDMS base.
    \item Adding a thin black silicone layer ($\leq0.5$ mm) for surface coating, which is made of the same silicone mixed with a black colorant (Silc Pig from Smooth-On).
    \item Installing a white LED ring (LUXEON 2835 4000K SMD LED) and a USB camera ($120^{\circ}$ FOV, $640\times480$ resolution) for illumination and image capture.
\end{enumerate}

The sensor operates at 30~Hz with depth inference, with dimensions of $29\times 29 \times 27$~mm and a total cost of approximately \$15. 

\begin{figure}[t!]
    \centering
    \includegraphics[width=0.95\linewidth]{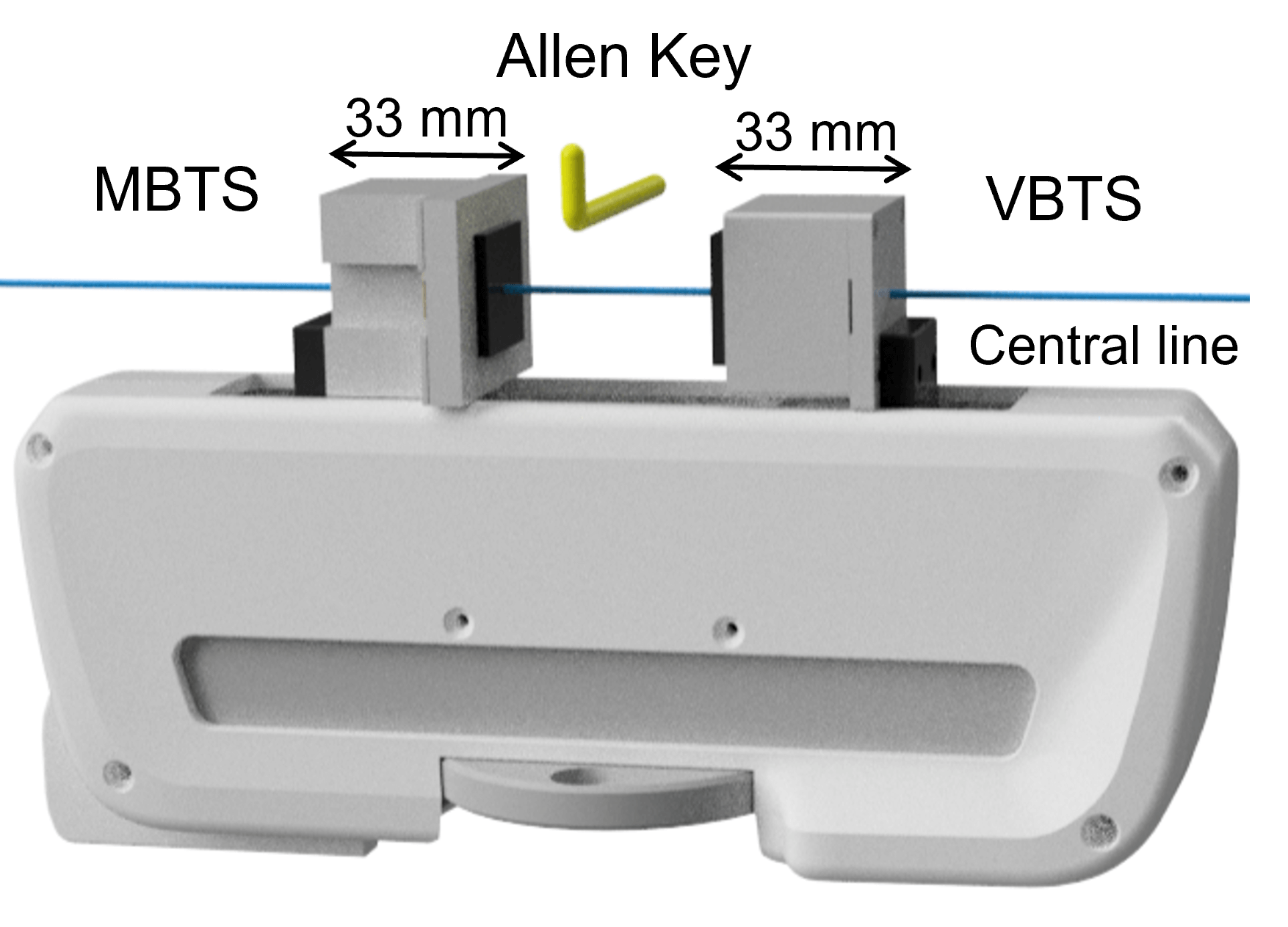}
    \caption{Data collection setup of \name.}
    \label{fig:calisetup}
\end{figure}

\subsection{Data Collection}
\label{sec:data-collect}
The preliminary setup for \ac{vbts}-guided super-resolution of \ac{mbts} requires identical contact conditions for both sensors. This involves two key aspects: 1) the contact modules of both sensors must have identical materials and dimensions, as confirmed earlier; and 2) the contact position and force must match during data collection.

To achieve this, we designed a symmetrically structured calibration setup, as shown in \cref{fig:calisetup}. The sensors are mounted on opposite sides of a gripper, with their contact modules aligned at the center. Images from the \ac{vbts} are calibrated, mirrored, and cropped to match the \ac{mbts} dimensions, with raw data ($325\times325$ pixels per frame) downsampled to $200\times200$ pixels for the dataset. A 3D-printed Allen key and the letter ``R'' are used to serve as the calibration object during training data collection. The gripper closes with a press force threshold of 10 N. The resulting dataset comprises 2025 pairs of tactile readings for the Allen key and 2025 pairs for the letter ``R''.

\section{Tactile Shape Reconstruction} \label{sec:recon}
Reconstructing the \ac{vbts} depth estimation $I$ from the \ac{mbts} reading $x$ can be inherently cross-modal. Specifically, multiple valid visual outcomes could correspond to similar magnetic measurements, due to surface texture, lighting, or sensor noise variations. 
A \ac{cvae} provides a probabilistic framework that is well-suited to handling such ambiguity. Unlike deterministic models, which map \(x\) to a single point estimate of \(I\), the \ac{cvae} models the distribution of possible outputs given \(x\). This capability allows the model to capture the complex relationships and variations in the tactile data, offering richer predictions. Moreover, by conditioning the generative process on \(x\), the \ac{cvae} ensures that the latent representations align with specific aspects of the magnetic reading, leading to a more interpretable and controllable generation process compared to standard (unconditional) \acp{vae}. 
In practice, a \ac{cvae} generally has a simpler network and inference procedure compared to \ac{gan} or diffusion models. This computational efficiency makes the \ac{cvae} more suitable for high-frequency operation in our scenarios, where fast inference from \ac{mbts} data is crucial. 

\subsection{Problem Formulation}
We consider the \ac{mbts} readings as a condition \(x\), which are typically vectors or spatially organized signals from magnetometers. Our goal is to learn a mapping \(x \mapsto I\), where \(I\) represents the corresponding reconstructed depth image from \ac{vbts}. 
Formally, we aim to learn the conditional distribution \(p_\theta(I \mid x)\). In a CVAE, this distribution is represented using a latent variable \(z\), such that:
\begin{equation}
    p_\theta(I \mid x) = \int p_\theta(I \mid z, x) \, p(z \mid x) \, dz.
    \label{eqn:problem}
\end{equation}
Since the true posterior \(p_\theta(z \mid x, I)\) is typically intractable, we introduce a variational approximation \(q_\phi(z \mid x, I)\), where \(\theta\) and \(\phi\) are the learnable parameters of the decoder and encoder networks, respectively.

To train the CVAE, we maximize the evidence lower bound (ELBO) on the log-likelihood of \(I\) given \(x\). The ELBO is given by:
\begin{equation}
\begin{split}
 \mathcal{L}(\theta, \phi; x, I)   = & \mathbb{E}_{q_\phi(z \mid x, I)} \bigl[ \log p_\theta(I \mid z, x) \bigr] \\
& - D_{\mathrm{KL}}\bigl(q_\phi(z \mid x, I) \,\|\, p(z \mid x)\bigr),
\end{split}
\label{eqn:loss}
\end{equation}
where \(D_{\mathrm{KL}}\) is the Kullback–Leibler divergence measuring how far \(q_\phi(z \mid x, I)\) deviates from the prior \(p(z \mid x)\). The first term encourages the decoder \(p_\theta(I \mid z, x)\) to reconstruct \(I\) accurately from \(z\) and \(x\), while the second term regularizes the encoder \(q_\phi(z \mid x, I)\) to remain close to the prior distribution \(p(z \mid x)\). During backpropagation, the reparameterization trick is employed to enable gradient-based optimization through the sampling process of the latent variable \(z\).

\begin{figure}[t!]
    \centering
    \includegraphics[width=0.95\linewidth]{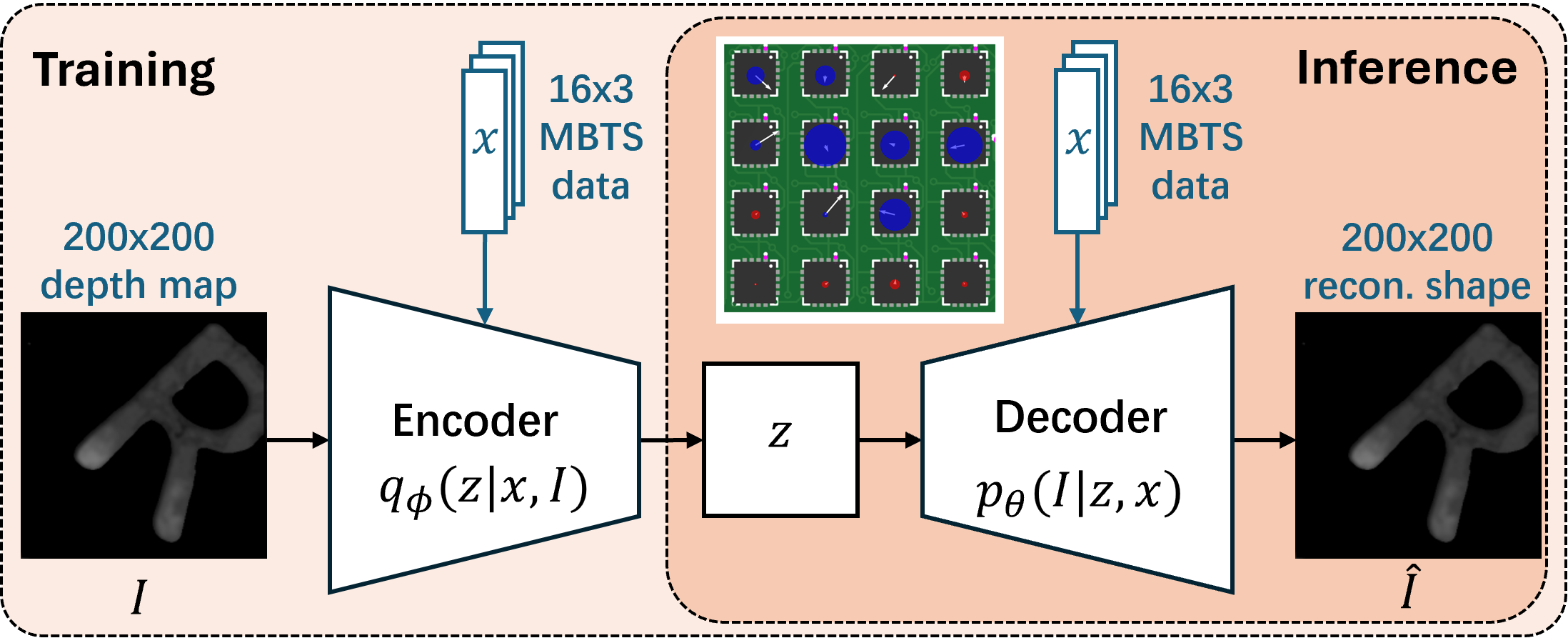}
    \caption{Network architecture of \name.}
    \label{fig:architecture}
\end{figure}

\subsection{Network Architecture and Implementation}
Our network architecture of \name is illustrated in \cref{fig:architecture}. The encoder network \(q_\phi(z \mid x, I)\) and decoder network \(p_\theta(I \mid z, x)\) are parameterized by \(\phi\) and \(\theta\), respectively. The encoder processes the visual reading \(I\) through convolutional and \ac{fc} layers, with the first layer incorporating a residual connection to produce the parameters of the approximate posterior distribution. The condition, \ie, the \ac{mbts} reading \(x\), is embedded via separate \ac{fc} layers and integrated into the encoder network as multiplicative conditioning at two stages to enhance memory retention and prevent catastrophic forgetting. The decoder takes the latent variable \(z\), sampled using the reparametrization trick, and generates a reconstruction \(\hat{I}\) through convolutional and \ac{fc} layers. Similar multiplicative conditioning with \(x\) is applied to the decoder. We model \(p_\theta(I \mid z, x)\) as a Gaussian distribution, with its mean and variance learned by the decoder, producing the reconstructed image \(\hat{I}\).

For training, we optimize the negative ELBO (defined in \cref{eqn:loss}) using mini-batch stochastic gradient descent. To stabilize training, we employ the reparameterization trick, sampling \(\epsilon\) from a standard Gaussian and computing \(z = \mu + \sigma \odot \epsilon\). Standard techniques such as batch/group normalization and max/average pooling are used to enhance network robustness and generalization. The number of latent dimensions is chosen to balance reconstruction accuracy and generative diversity. After training, given a new \ac{mbts} reading \(x\), we sample \(z\) from the prior \(p_\theta(z \mid x)\) and generate multiple plausible predictions \(\hat{I}\), capturing the inherent uncertainty and variability in tactile sensing.

\section{Experiments}  
In this paper, two sets of experiments are conducted on a range of novel objects to evaluate: 1)~the quality of shape reconstruction across diverse seen/unseen partial/full objects; 2)~the real-world applicability of \name in object reorientation tasks. 

\subsection{Shape Reconstruction}  
\textbf{Data Preparation:}  
From the collected dataset (see \cref{sec:data-collect}), the Allen key and letter ``R'' are selected for training due to their distinct shape characteristics. \ac{vbts} data are de-noised and down-sampled from $325 \times 325$ to $200 \times 200$ pixels to balance image quality and training efficiency. \ac{mbts} data are clamped to $\pm500$ and normalized to $[-1, 1]$ for training stability.  

\textbf{Network Details:}  
The lightweight \name architecture comprises 47.2M parameters. The encoder consists of 6 \ac{cnn} blocks with 2 MaxPooling layers and 1 AvgPooling layer before 2 final \ac{fc} layers (latent dimension: 512). All encoder blocks use BatchNorm and GELU activation. The decoder includes 4 ConvT blocks with 8 intermediate \ac{cnn} blocks, employing GroupNorm and ReLU in ConvT blocks, and BatchNorm and GELU in \ac{cnn} blocks. Conditions are embedded via 2 \ac{fc} layers (GELU-activated) in both the encoder and decoder. Training is performed on an NVIDIA RTX 3090 GPU for approximately 350 epochs using a training set containing 1822 (Allen key) and 1822 (letter ``R'') samples (3645 in total), with a fixed learning rate of $1\text{e-}4$ using the AdamW optimizer. Shape reconstruction from a single \ac{mbts} reading completes within 2.5~ms on the same hardware, enabling a runtime frequency of approximately 95~Hz, including an 8~ms sensor latency and communication overhead.

\textbf{Baselines:}
We evaluate \name against three baselines to assess its tactile shape reconstruction capabilities: 1)~Bilinear/Bicubic interpolation of z-axis \ac{mbts} data, serving as naive spatial upsampling baselines; 2)~\name (z-axis), which uses only z-axis magnetic data for reconstruction; and 3)~\name (AK), trained solely on Allen key data (half of the training set) to analyze generalization limits. \name employs full three-axis \ac{mbts} data and multi-object training (Allen key + letter ``R''). All methods are tested on unseen objects to evaluate generalization.  

\textbf{Metrics:}
To evaluate the performance of \name on tactile shape reconstruction, we compare different models and versions of \name across three metrics on a set of unseen objects (see \cref{fig:gallery}): \ac{psnr}~\cite{wang2020deep}: Measures pixel-level accuracy per image. Values below 20~dB indicate severe defects and are unacceptable. \ac{ssim}~\cite{wang2004image}: Evaluates human-like visual quality per image, considering luminance, contrast, and texture. Scores range from -1 to 1, with negative values being unacceptable and higher values indicating better quality. \ac{fid}~\cite{heusel2017gans}: Assesses the similarity between the reconstructed and ground truth visual images. Lower values indicate better alignment.  

\subsection{Shape Reconstruction Results}
As shown in \cref{tab:results}, \name outperforms all baselines across all metrics, demonstrating its efficacy in tactile shape reconstruction. Bilinear and Bicubic interpolation methods yield poor performance, confirming that naive upsampling fails to recover high-resolution tactile details. SuperMag (z-axis), using only single-axis magnetic data, achieves moderate improvements but remains limited by incomplete deformation information. SuperMag (AK), trained on a single object (Allen key), further reduces \ac{fid} and increases \ac{psnr}, highlighting the benefits of multi-axis magnetic conditioning.

\begin{table}[th!]
    \centering
    \caption{Comparison of SuperMag against baselines.}
    \resizebox{\linewidth}{!}{%
    \setlength{\tabcolsep}{8pt} 
    \setstretch{1.1} 
    \begin{tabular}{l>{\columncolor{blue!20}}c >{\columncolor{blue!20}}c >{\columncolor{blue!20}}c}
    \toprule
    \textbf{Method Name} & \textbf{FID}$\downarrow$ & \textbf{PSNR[dB]}$\uparrow$ & \textbf{SSIM}$\uparrow$ \\
    \midrule
    Bilinear & 402.63 & 8.03 $\pm$ 1.78 & 0.10 $\pm$ 0.04 \\
    Bicubic & 309.10 & 6.75 $\pm$ 1.60 & 0.10 $\pm$ 0.04 \\
    SuperMag (z-axis) & 234.16 & 20.86 $\pm$ 1.93 & 0.69 $\pm$ 0.08 \\
    SuperMag (AK) & 213.43 & 22.36 $\pm$ 2.32 & 0.65 $\pm$ 0.07 \\
    SuperMag (Ours) & \textbf{210.10} & \textbf{24.24}  $\pm$ 2.88& \textbf{0.78} $\pm$ 0.06\\
    \bottomrule
    \end{tabular}%
    }%
    \label{tab:results}
\end{table}

The full SuperMag, leveraging multi-axis data and multi-object training, achieves the best performance. This underscores the critical role of training diversity and cross-modal supervision. The superior \ac{ssim} indicates high perceptual fidelity in reconstructed shapes, while the \ac{psnr} validates pixel-level accuracy. The results demonstrate that integrating multi-axis \ac{mbts} data with \ac{vbts}-guided supervision effectively bridges the resolution-frequency trade-off, enabling real-time, high-resolution tactile perception for robots.  

\begin{figure*}[t]
    \centering
    \includegraphics[width=\linewidth]{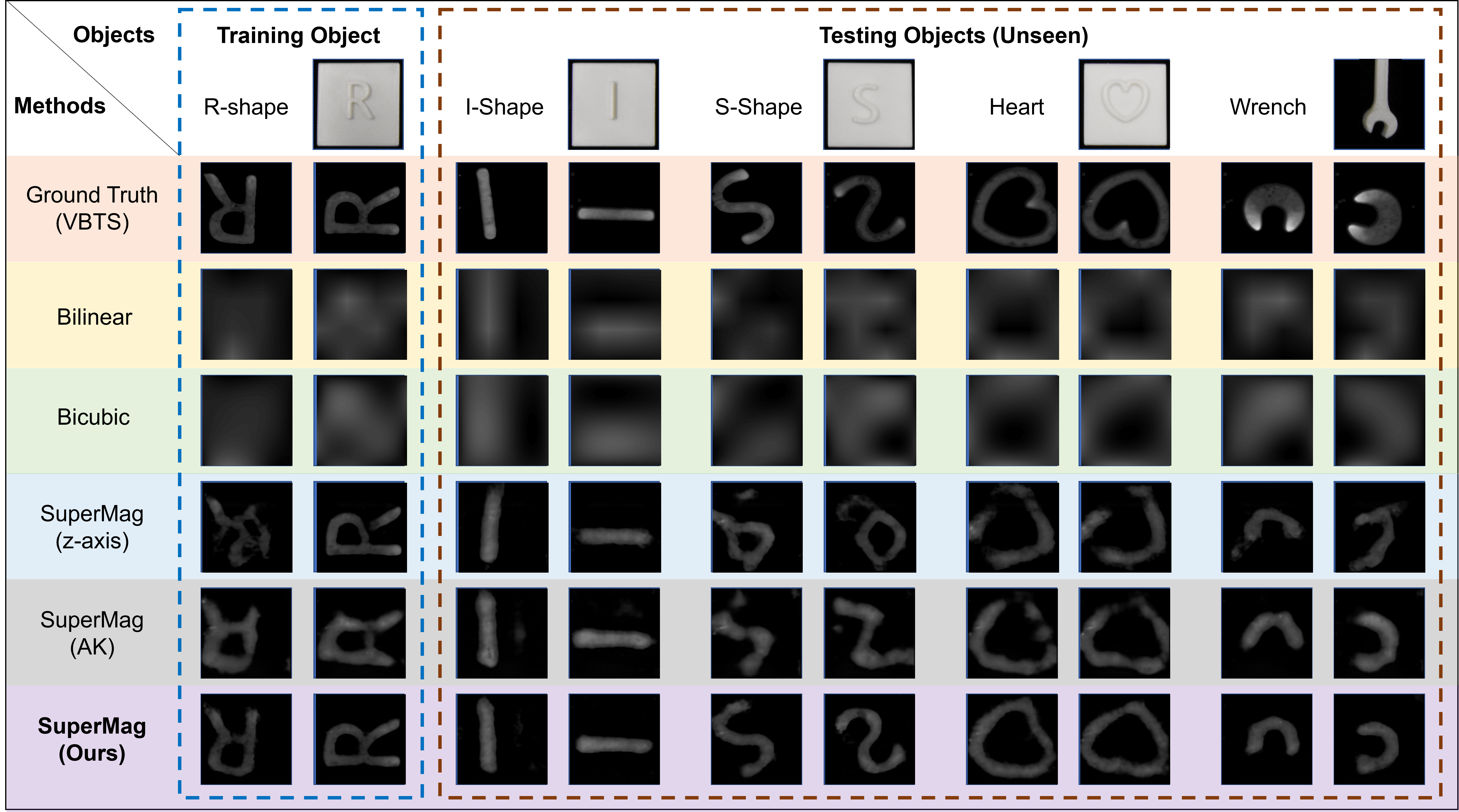}
    \caption{\textbf{Shape reconstruction results for ground truth, baselines, and \name on seen training object and unseen testing objects.}  
    The ground truth is from the vision-based tactile sensor (VBTS). Baselines are Bilinear and Bicubic interpolation of z-axis magnetic-based tactile sensor (MBTS) data. \name (z-axis) is trained on R-shape z-axis MBTS data; \name(AK) is trained on Allen key (AK) MBTS data, noted that R-shape is an unseen object for \name(AK); \name is trained on both Allen key and R-shape MBTS data.}
    \label{fig:gallery}
\end{figure*}

\cref{fig:gallery} provides a qualitative comparison of all methods, displaying the generated high-resolution images alongside the ground truth for visual evaluation.

\subsection{In-hand Pose Estimation for Object Reorientation}
\begin{figure}[ht!]
    \centering
    \includegraphics[width=0.9\linewidth]{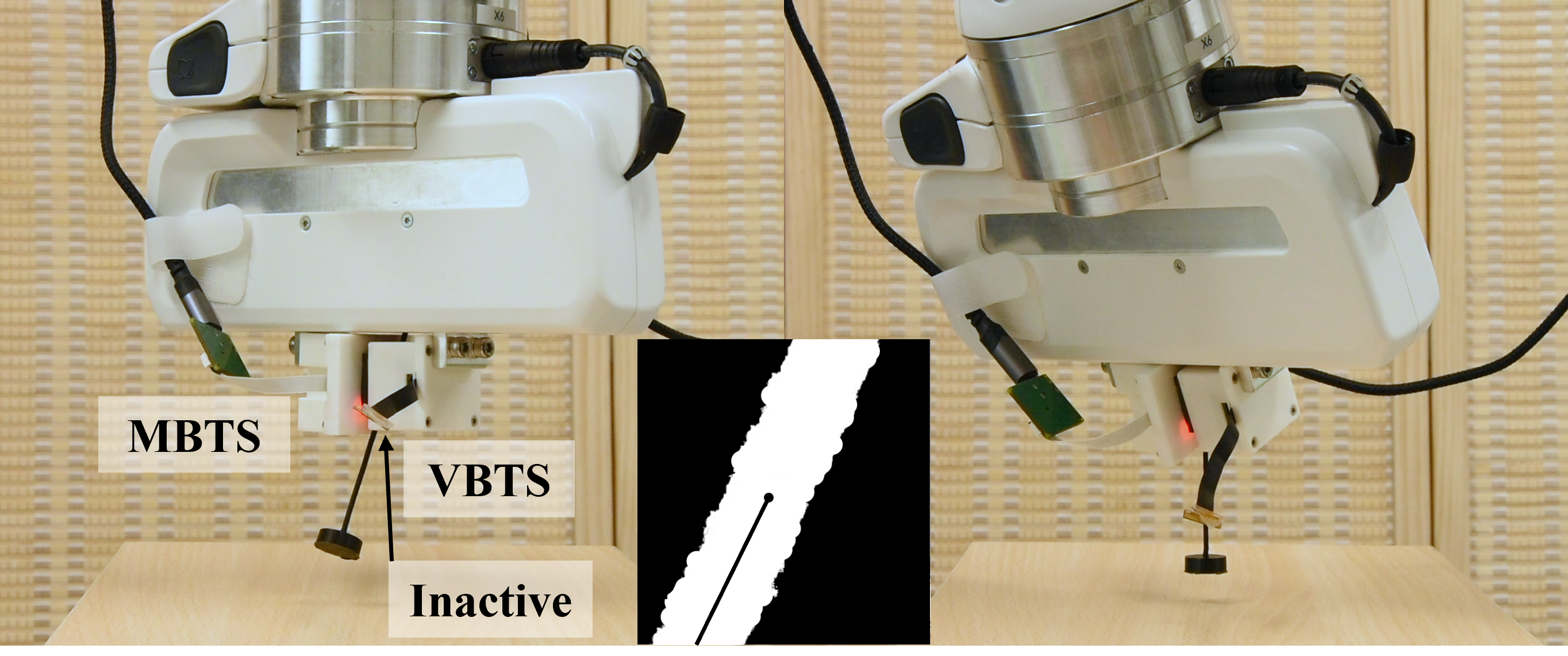}
    \caption{In-hand Pose Estimation for Object Reorientation.}
    \label{fig:exp}
\end{figure}
In this section, we demonstrate in-hand pose estimation and object reorientation tasks using tactile feedback to evaluate the impact of super-resolution on \ac{mbts}. The sensor is mounted on the gripper of a Franka Research 3 7-DoF robotic arm (see \cref{fig:exp}), with communication handled via USB for the sensors and Ethernet for the gripper and arm.

Three rod-shaped objects of varying sizes and geometries were tested, with two random orientations evaluated for each object. For \name, the generated images were preprocessed using outlier removal and Gaussian filtering, followed by PCA-based angle estimation. The baseline method, producing 4×4 resolution outputs, was upscaled to 16×16 via bilinear interpolation and processed similarly.  

The results demonstrate the superiority of the \name, achieving a 100\% success rate (6/6 trials) in angle correction, compared to the baseline's 16.7\% success rate (1/6 trials). The super-resolution capability enables the \ac{mbts} to estimate in-hand object poses with higher precision, unlocking capabilities that were previously infeasible.

\begin{figure}[th!]
    \centering
    \includegraphics[width=0.97\linewidth]{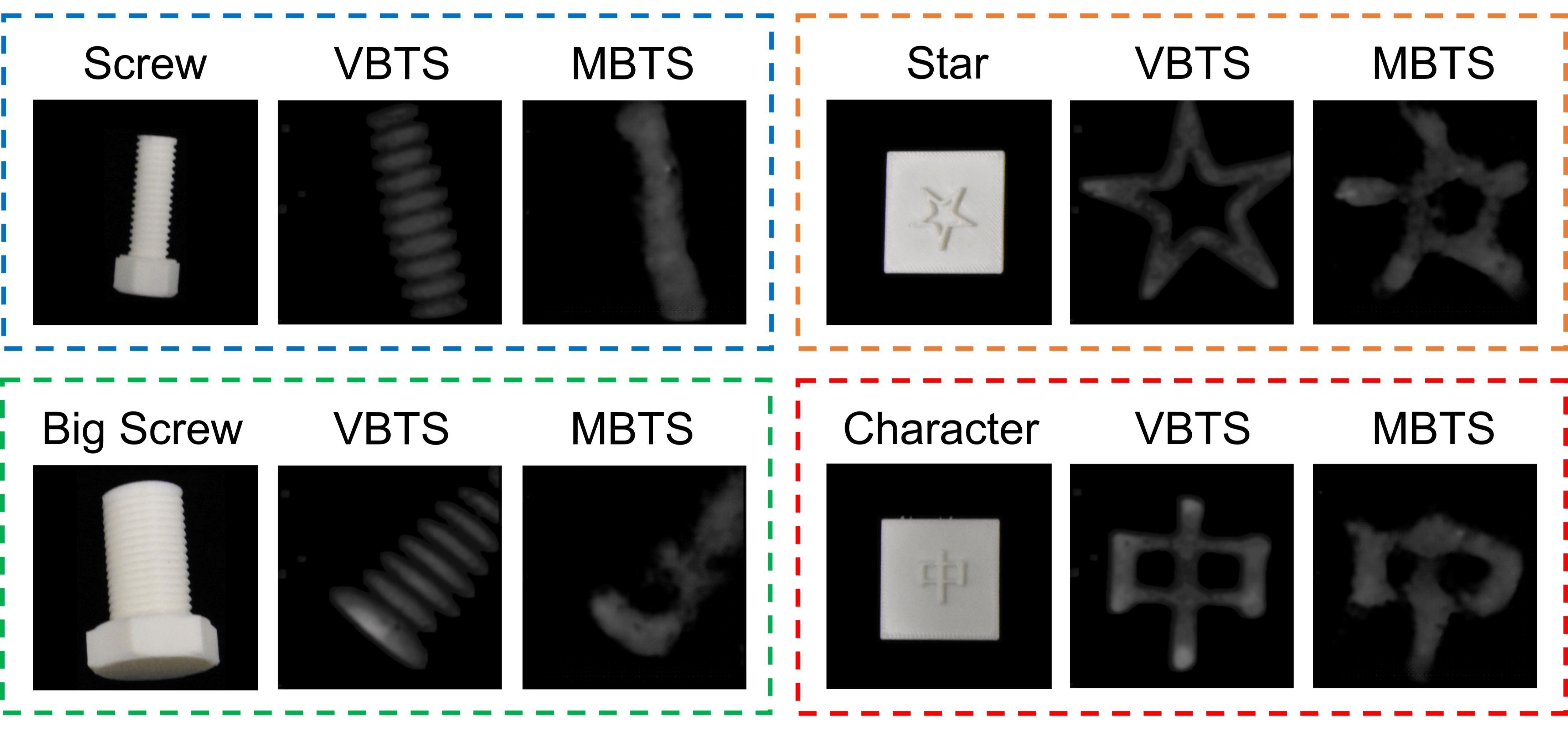}
    \caption{Example results of \name shape reconstruction for unseen objects with fine texture details.}
    \label{fig:limitation}
\end{figure}

\section{Discussion \& Conclusion}\label{sec:discussion}
This work presents \name, a tactile shape reconstruction method that enables super-resolution of \acf{mbts} using \acf{vbts} data. Leveraging co-designed sensors with identical contact modules and a symmetric calibration setup, we train a \acf{cvae} to infer high-resolution tactile shapes from low-resolution \ac{mbts} inputs. \name reconstructs $200\times200$ pixel shapes from $4\times4\times3$ taxel arrays, outperforming baselines in \ac{fid}, \ac{psnr}, and \ac{ssim}, while operating at a 95~Hz. We further demonstrate its utility in an object reorientation task via pose estimation.

However, several limitations and future directions remain. First, the proposed method is currently constrained to \ac{mbts} sensors equipped with contact modules that match the dimensions and silicone material of the \ac{vbts}. Additionally, the use of \ac{mbts} may be unsuitable for grasping magnetizable materials such as steel. Future work will investigate the transferability and generalization of the approach across a broader range of taxel-based sensors with varying structural and material designs. 
Second, while \name excels at shape contour reconstruction, its ability to recover fine details remains inferior to \ac{vbts} (see \cref{fig:limitation}), requiring further refinement. 
Finally, the inherent limitation of \ac{vbts} in detecting large planar surfaces impacts \ac{mbts} performance, necessitating additional research. These advancements aim to further bridge the gap between high-frequency and high-resolution tactile sensing for robotics.

\clearpage
\setstretch{0.995}
\bibliographystyle{ieeetr}
\bibliography{IEEEfull}

\begin{thebibliography}{10}

\bibitem{cutkosky2008force}
M.~R. Cutkosky, R.~D. Howe, and W.~R. Provancher, ``Force and tactile sensors,'' {\em Springer Handbook of Robotics}, pp.~455--476, 2008.

\bibitem{yousef2011tactile}
H.~Yousef, M.~Boukallel, and K.~Althoefer, ``Tactile sensing for dexterous in-hand manipulation in robotics—a review,'' {\em Sensors and Actuators A: physical}, vol.~167, no.~2, pp.~171--187, 2011.

\bibitem{dawood2023learning}
A.~B. Dawood, C.~Coppola, and K.~Althoefer, ``Learning decoupled multi-touch force estimation, localization and stretch for soft capacitive e-skin,'' in {\em IEEE International Conference on Robotics and Automation (ICRA)}, pp.~614--619, IEEE, 2023.

\bibitem{zhu2022recent}
Y.~Zhu, Y.~Liu, Y.~Sun, Y.~Zhang, and G.~Ding, ``Recent advances in resistive sensor technology for tactile perception: A review,'' {\em IEEE Sensors Journal}, vol.~22, no.~16, pp.~15635--15649, 2022.

\bibitem{stassi2014flexible}
S.~Stassi, V.~Cauda, G.~Canavese, and C.~F. Pirri, ``Flexible tactile sensing based on piezoresistive composites: A review,'' {\em Sensors}, vol.~14, no.~3, pp.~5296--5332, 2014.

\bibitem{lin2021skin}
W.~Lin, B.~Wang, G.~Peng, Y.~Shan, H.~Hu, and Z.~Yang, ``Skin-inspired piezoelectric tactile sensor array with crosstalk-free row+ column electrodes for spatiotemporally distinguishing diverse stimuli,'' {\em Advanced Science}, vol.~8, no.~3, p.~2002817, 2021.

\bibitem{qu2022fingerprint}
X.~Qu, J.~Xue, Y.~Liu, W.~Rao, Z.~Liu, and Z.~Li, ``Fingerprint-shaped triboelectric tactile sensor,'' {\em Nano Energy}, vol.~98, p.~107324, 2022.

\bibitem{hou2024location}
J.~Hou, X.~Zhou, and A.~J. Spiers, ``Location and orientation super-resolution sensing with a cost-efficient and repairable barometric tactile sensor,'' {\em IEEE Transactions on Robotics (T-RO)}, 2024.

\bibitem{yao2024recent}
N.~Yao and S.~Wang, ``Recent progress of optical tactile sensors: A review,'' {\em Optics \& Laser Technology}, vol.~176, p.~111040, 2024.

\bibitem{man2022recent}
J.~Man, G.~Chen, and J.~Chen, ``Recent progress of biomimetic tactile sensing technology based on magnetic sensors,'' {\em Biosensors}, vol.~12, no.~11, p.~1054, 2022.

\bibitem{wang2016design}
H.~Wang, G.~De~Boer, J.~Kow, A.~Alazmani, M.~Ghajari, R.~Hewson, and P.~Culmer, ``Design methodology for magnetic field-based soft tri-axis tactile sensors,'' {\em Sensors}, vol.~16, no.~9, p.~1356, 2016.

\bibitem{sun2022guiding}
H.~Sun and G.~Martius, ``Guiding the design of superresolution tactile skins with taxel value isolines theory,'' {\em Science Robotics}, vol.~7, no.~63, p.~eabm0608, 2022.

\bibitem{huang2024multipole}
J.~Huang, Y.~Lou, X.~Xiong, X.~Yang, and Y.~Li, ``Multipole magnetic tactile sensor with super-resolution for 3d pose estimation,'' {\em IEEE Transactions on Instrumentation and Measurement (TIM)}, 2024.

\bibitem{yan2021soft}
Y.~Yan, Z.~Hu, Z.~Yang, W.~Yuan, C.~Song, J.~Pan, and Y.~Shen, ``Soft magnetic skin for super-resolution tactile sensing with force self-decoupling,'' {\em Science Robotics}, vol.~6, no.~51, p.~eabc8801, 2021.

\bibitem{hu2024large}
H.~Hu, C.~Zhang, X.~Lai, H.~Dai, C.~Pan, H.~Sun, D.~Tang, Z.~Hu, J.~Fu, T.~Li, {\em et~al.}, ``Large-area magnetic skin for multi-point and multi-scale tactile sensing with super-resolution,'' {\em npj Flexible Electronics}, vol.~8, no.~1, p.~42, 2024.

\bibitem{wu2022tactile}
B.~Wu, Q.~Liu, and Q.~Zhang, ``Tactile pattern super resolution with taxel-based sensors,'' in {\em IEEE/RSJ International Conference on Intelligent Robots and Systems (IROS)}, pp.~3644--3650, IEEE, 2022.

\bibitem{shah2021design}
U.~H. Shah, R.~Muthusamy, D.~Gan, Y.~Zweiri, and L.~Seneviratne, ``On the design and development of vision-based tactile sensors,'' {\em Journal of Intelligent \& Robotic Systems}, vol.~102, pp.~1--27, 2021.

\bibitem{zhong2023touching}
S.~Zhong, A.~Albini, O.~P. Jones, P.~Maiolino, and I.~Posner, ``Touching a nerf: Leveraging neural radiance fields for tactile sensory data generation,'' in {\em Annual Conference on Robot Learning (CoRL)}, pp.~1618--1628, PMLR, 2023.

\bibitem{li2024vision}
S.~Li, Z.~Wang, C.~Wu, X.~Li, S.~Luo, B.~Fang, F.~Sun, X.-P. Zhang, and W.~Ding, ``When vision meets touch: A contemporary review for visuotactile sensors from the signal processing perspective,'' {\em IEEE Journal of Selected Topics in Signal Processing}, 2024.

\bibitem{bhirangi2021reskin}
R.~Bhirangi, T.~Hellebrekers, C.~Majidi, and A.~Gupta, ``Reskin: versatile, replaceable, lasting tactile skins,'' in {\em Annual Conference on Robot Learning (CoRL)}, 2021.

\bibitem{lin2023dtact}
C.~Lin, Z.~Lin, S.~Wang, and H.~Xu, ``Dtact: A vision-based tactile sensor that measures high-resolution 3d geometry directly from darkness,'' in {\em IEEE International Conference on Robotics and Automation (ICRA)}, pp.~10359--10366, IEEE, 2023.

\bibitem{rehan2022soft}
M.~Rehan, M.~M. Saleem, M.~I. Tiwana, R.~I. Shakoor, and R.~Cheung, ``A soft multi-axis high force range magnetic tactile sensor for force feedback in robotic surgical systems,'' {\em Sensors}, vol.~22, no.~9, p.~3500, 2022.

\bibitem{ge2019bimodal}
J.~Ge, X.~Wang, M.~Drack, O.~Volkov, M.~Liang, G.~S. Ca{\~n}{\'o}n~Berm{\'u}dez, R.~Illing, C.~Wang, S.~Zhou, J.~Fassbender, {\em et~al.}, ``A bimodal soft electronic skin for tactile and touchless interaction in real time,'' {\em Nature communications}, vol.~10, no.~1, p.~4405, 2019.

\bibitem{dai2024split}
H.~Dai, C.~Zhang, C.~Pan, H.~Hu, K.~Ji, H.~Sun, C.~Lyu, D.~Tang, T.~Li, J.~Fu, {\em et~al.}, ``Split-type magnetic soft tactile sensor with 3d force decoupling,'' {\em Advanced Materials}, vol.~36, no.~11, p.~2310145, 2024.

\bibitem{li2024tactile}
J.~Li, H.~Qin, Z.~Song, L.~Hou, and H.~Li, ``A tactile sensor based on magnetic sensing: Design and mechanism,'' {\em IEEE Transactions on Instrumentation and Measurement (TIM)}, 2024.

\bibitem{gribbon2004novel}
K.~T. Gribbon and D.~G. Bailey, ``A novel approach to real-time bilinear interpolation,'' in {\em Proceedings. DELTA 2004. Second IEEE international workshop on electronic design, test and applications}, pp.~126--131, IEEE, 2004.

\bibitem{hu2022wireless}
H.~Hu, C.~Zhang, C.~Pan, H.~Dai, H.~Sun, Y.~Pan, X.~Lai, C.~Lyu, D.~Tang, J.~Fu, {\em et~al.}, ``Wireless flexible magnetic tactile sensor with super-resolution in large-areas,'' {\em ACS nano}, vol.~16, no.~11, pp.~19271--19280, 2022.

\bibitem{lepora2015superresolution}
N.~F. Lepora and B.~Ward-Cherrier, ``Superresolution with an optical tactile sensor,'' in {\em IEEE/RSJ International Conference on Intelligent Robots and Systems (IROS)}, pp.~2686--2691, IEEE, 2015.

\bibitem{yamaguchi2017implementing}
A.~Yamaguchi and C.~G. Atkeson, ``Implementing tactile behaviors using fingervision,'' in {\em IEEE/RAS International Conference on Humanoid Robotics (Humanoids)}, pp.~241--248, IEEE, 2017.

\bibitem{cui2021self}
S.~Cui, R.~Wang, J.~Hu, C.~Zhang, L.~Chen, and S.~Wang, ``Self-supervised contact geometry learning by gelstereo visuotactile sensing,'' {\em IEEE Transactions on Instrumentation and Measurement (TIM)}, vol.~71, pp.~1--9, 2021.

\bibitem{ward2018tactip}
B.~Ward-Cherrier, N.~Pestell, L.~Cramphorn, B.~Winstone, M.~E. Giannaccini, J.~Rossiter, and N.~F. Lepora, ``The tactip family: Soft optical tactile sensors with 3d-printed biomimetic morphologies,'' {\em Soft robotics}, vol.~5, no.~2, pp.~216--227, 2018.

\bibitem{sato2009finger}
K.~Sato, K.~Kamiyama, N.~Kawakami, and S.~Tachi, ``Finger-shaped gelforce: sensor for measuring surface traction fields for robotic hand,'' {\em IEEE Transactions on Haptics (ToH)}, vol.~3, no.~1, pp.~37--47, 2009.

\bibitem{wang2024large}
M.~Wang, W.~Li, H.~Liang, B.~Li, K.~Althoefer, Y.~Su, and H.~Liu, ``Large-scale deployment of vision-based tactile sensors on multi-fingered grippers,'' in {\em IEEE/RSJ International Conference on Intelligent Robots and Systems (IROS)}, pp.~13946--13952, IEEE, 2024.

\bibitem{li2024minitac}
W.~Li, Z.~Zhao, L.~Cui, W.~Zhang, H.~Liu, L.-A. Li, and Y.~Zhu, ``Minitac: An ultra-compact 8 mm vision-based tactile sensor for enhanced palpation in robot-assisted minimally invasive surgery,'' {\em Robotics and Automation Letters (RA-L)}, 2024.

\bibitem{yuan2017gelsight}
W.~Yuan, S.~Dong, and E.~H. Adelson, ``Gelsight: High-resolution robot tactile sensors for estimating geometry and force,'' {\em Sensors}, vol.~17, no.~12, p.~2762, 2017.

\bibitem{donlon2018gelslim}
E.~Donlon, S.~Dong, M.~Liu, J.~Li, E.~Adelson, and A.~Rodriguez, ``Gelslim: A high-resolution, compact, robust, and calibrated tactile-sensing finger,'' in {\em IEEE/RSJ International Conference on Intelligent Robots and Systems (IROS)}, pp.~1927--1934, IEEE, 2018.

\bibitem{lambeta2020digit}
M.~Lambeta, P.-W. Chou, S.~Tian, B.~Yang, B.~Maloon, V.~R. Most, D.~Stroud, R.~Santos, A.~Byagowi, G.~Kammerer, {\em et~al.}, ``Digit: A novel design for a low-cost compact high-resolution tactile sensor with application to in-hand manipulation,'' {\em Robotics and Automation Letters (RA-L)}, vol.~5, no.~3, pp.~3838--3845, 2020.

\bibitem{mckenney1995fast}
A.~McKenney, L.~Greengard, and A.~Mayo, ``A fast poisson solver for complex geometries,'' {\em Journal of Computational Physics}, vol.~118, no.~2, pp.~348--355, 1995.

\bibitem{yuan2015measurement}
W.~Yuan, R.~Li, M.~A. Srinivasan, and E.~H. Adelson, ``Measurement of shear and slip with a gelsight tactile sensor,'' in {\em IEEE International Conference on Robotics and Automation (ICRA)}, pp.~304--311, IEEE, 2015.

\bibitem{li20233}
W.~Li, M.~Wang, J.~Li, Y.~Su, D.~K. Jha, X.~Qian, K.~Althoefer, and H.~Liu, ``L3f-touch: A wireless gelsight with decoupled tactile and three-axis force sensing,'' {\em Robotics and Automation Letters (RA-L)}, 2023.

\bibitem{zhao2024tac}
Z.~Zhao, Y.~Li, W.~Li, Z.~Qi, L.~Ruan, Y.~Zhu, and K.~Althoefer, ``Tac-man: Tactile-informed prior-free manipulation of articulated objects,'' {\em IEEE Transactions on Robotics (T-RO)}, 2024.

\bibitem{wang2020deep}
Z.~Wang, J.~Chen, and S.~C. Hoi, ``Deep learning for image super-resolution: A survey,'' {\em Transactions on Pattern Analysis and Machine Intelligence (TPAMI)}, vol.~43, no.~10, pp.~3365--3387, 2020.

\bibitem{wang2004image}
Z.~Wang, A.~C. Bovik, H.~R. Sheikh, and E.~P. Simoncelli, ``Image quality assessment: from error visibility to structural similarity,'' {\em Transactions on Image Processing (TIP)}, vol.~13, no.~4, pp.~600--612, 2004.

\bibitem{heusel2017gans}
M.~Heusel, H.~Ramsauer, T.~Unterthiner, B.~Nessler, and S.~Hochreiter, ``Gans trained by a two time-scale update rule converge to a local nash equilibrium,'' {\em Advances in Neural Information Processing Systems (NIPS)}, vol.~30, 2017.

\end{thebibliography}

\end{document}